\documentclass[letterpaper, 10pt, conference]{ieeeconf}
\IEEEoverridecommandlockouts
\usepackage{cite}
\usepackage{graphicx}
\usepackage{amsmath,amssymb}
\usepackage{url}
\usepackage{balance}

\usepackage{graphicx}

\title{\LARGE \bf Task-Critical Part Alignment Improves Imitation of Object Interactions}

\author{Denis Hadjivelichkov,
        Sebastian {\AE}gidius,
        Valerio Modugno,
        Dimitrios Kanoulas
\thanks{The authors are with University College London, 66 Gower Street, WC1E 6BT, London, UK. Dimitrios Kanoulas is also with Archimedes/Athena RC, Greece. Corresponding author is Denis Hadjivelichkov. }
\thanks{This work was supported by UKRI FLF [MR/V025333/1] and CDT of FAI [EP/S021566/1]. For Open Access, the author has applied a CC BY copyright license to any manuscript version arising from this submission.
}%
}

\begin{document}
\maketitle
\thispagestyle{empty}
\pagestyle{empty}

\begin{abstract}

This paper presents enhancements to the DINOBot framework for imitation learning of object-object interactions. We introduce two key contributions aimed at improving the accuracy and efficiency of alignment in imitation-based robotic tasks. First, we integrate RANdom SAmple Consensus (RANSAC) for outlier rejection during the alignment phase, effectively mitigating the impact of noisy depth data and spurious feature matches. This improves robustness when transferring demonstrations to real-world applications. Second, we refine the semantic correspondence process by focusing on task-relevant interaction regions, ensuring that the learned representation prioritizes critical object features that influence successful manipulation. We evaluated our approach in a range of real-world manipulation tasks, including picking, insertion, shaking, and pouring, using a robotic system. The experimental results demonstrate an overall improvement of 13.8\% in task success rates compared to the baseline, along with faster convergence in the alignment process. These results highlight the effectiveness of our enhancements in improving precision and efficiency for object-object interaction imitation in robotics.
\end{abstract}

\section{Introduction}

In robotic manipulation, transferring behaviors acquired from minimal demonstrations to previously unseen object instances remains an open problem. Specifically, a critical challenge is the generalization of manipulation skills across objects with diverse shapes, sizes, and semantic attributes. In real-world scenarios, robots must not only identify objects, but also infer the underlying spatial and functional relationships among them. For example, effective alignment of a tool with its target requires an appreciation of both the overall geometry and the localized, task-critical regions of the objects involved.

Existing approaches in imitation learning have largely focused on end-to-end visual imitation learning. Recent object-centric approaches such as DINOBot~\cite{dipalo2024dinobot} have shown that sample efficiency can be significantly improved by focusing only on object interactions, although limited to object interaction skills. However, global object alignment strategies may overlook the fine-grained details essential for successful execution, particularly in tasks where only specific regions (e.g., the tip of a screwdriver or the handle of a mug) are functionally relevant.  Given the promising improvements in object part correspondence~\cite{Hadjivelichkov2022affcorrs} and pose alignment demonstrated in~\cite{Tekden2022, Hadjivelichkov2025}, our objective is to determine how imitation learning frameworks such as DINOBot can be modified to focus on the task-critical parts and whether this leads to significant benefits.

In this work, we introduce several key innovations. First, we improve the automatic detection of task-critical Regions of Interaction (RoIs) introduced in~\cite{Hadjivelichkov2025} by considering all pair of object points that are bidirectionally closest. Secondly, we propose an interaction-focused correspondence strategy that focuses on RoIs. By confining semantic matching to the task-critical parts of the objects, our method effectively reduces ambiguity and improves alignment precision. Finally, we integrate RANSAC (Random Sample Consensus) into the alignment phase to robustly reject outliers resulting from depth noise and inaccurate correspondences, thereby ensuring that only consistent matches contribute to the transformation estimation.

Our experimental evaluation on a set of real-world manipulation tasks demonstrates that these enhancements yield a 13.8\% improvement in the overall success rate of the tasks compared to the baseline DINOBot system. Furthermore, our approach reduces the number of alignment iterations required for convergence, thereby accelerating the imitation process. These improvements in robustness and efficiency pave the way for complex robotic applications in industrial automation, logistics, and assistive robotics, where both adaptability and precision are important.
\begin{figure}[!t]
    \centering
    \includegraphics[width=\linewidth]{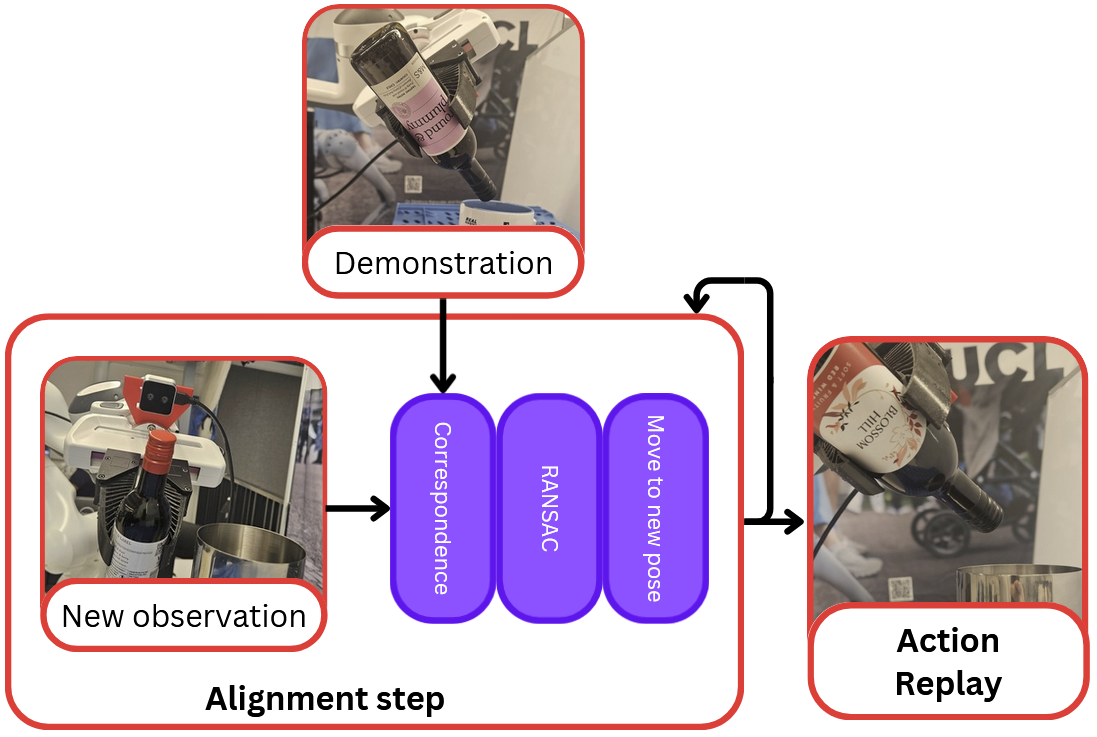}
    \caption{Overview of the method: alignment steps are iteratively performed until alignment is complete, followed by a replay phase. Our innovations lay in the Correspondence and RANSAC parts of the alignment phase.}
    \label{fig:hook}
\end{figure}
\section{Related Work}
Learning manipulation policies from offline visual demonstrations has gained rapid improvements following the advent of large vision models. Prior works~\cite{zeng2020transporter,pmlr-v164-shridhar22a} learned simple 2D planar
pick-and-place tasks. Recently, some works have shown impressive results in the 3D manipulation of objects in environments with significant novelties~\cite{pmlr-v270-zhao25b,Bauza2024,pmlr-v229-stone23a,lin2024datascalinglawsimitation,shridhar2022peract}. Interestingly, the benefits of increasing the number of task demonstrations reach a plateau, indicating that the diversity of demonstrations, that is, the capture of a range of different environments or task configurations, is more crucial than the size of the dataset~\cite{lin2024datascalinglawsimitation}. This insight is particularly relevant for few-shot imitation learning, where the ability to adapt to novel environments with minimal data is critical.

One-shot imitation learning aims to learn skills from a single new demonstration trajectory per task, effectively generalizing to novel variations of that tasks without further training~\cite{Mandi2022, pmlr-v155-dasari21a,pmlr-v78-finn17a,felice2023}. Several works present one-shot imitation learning methods for skills using a known object~\cite{vitiello2023one, johns2021coarse-to-fine,Biza2023}. The methods are capable of achieving fine-grained manipulation tasks by simply estimating the relative pose transformation from the object to the end effector at the start of the provided demonstration and aligning it in a novel arrangement. Following this alignment, the demonstration trajectory can simply be transferred relative to its new initial pose, by either transforming a Cartesian pose trajectory, or replaying the demonstrated Cartesian velocities.  Valassakis et al.~\cite{valassakis2022dome} demonstrate with DOME that the same paradigm can be successfully applied in sim-to-real transfer of manipulation skills.
Papagianis et al.~\cite{papagiannis2024adapting} highlight the importance of adapting the end-effector pose during deployment with respect to the tool object's current pose, allowing a more accurate trajectory replay. Norman Di Palo et al.~\cite{dipalo2024on} further extend this family of methods to novel object instances -- this is done by first retrieving a relevant demonstration and then aligning with the demonstrated manipulandum with the observed one using a trained alignment policy. Finally, DINOBot~\cite{dipalo2024dinobot} replaces the learned alignment with an alignment based on semantic point correspondences instead, significantly reducing the complexity of the method.

Although effective in many cases, global alignment methods can neglect task-critical details. Recent works~\cite{Tekden2022,Hadjivelichkov2025} have argued for the importance of focusing on the functional regions of objects -- TAX-Pose~\cite{pan2022taxpose}, a cross-pose correspondence model has shown improved performance when functional task-critical object parts are used instead of full objects~\cite{Hadjivelichkov2025}.

\section{Method}
In this section, we investigate the failure modes of the DINOBot framework and propose a novel approach to address its limitations during the tool alignment phase. In particular, we propose how DINOBot could be adapted to using only task-centric features.

\subsection{Problem Setup}
Given a single example demonstration of an object manipulation skill, we want to find and manipulate a novel, but similar object; for example, one that can afford a similar skill. The demonstration consists of a pose trajectory and its corresponding set of observations from a wrist-mounted camera. In this work, the point sets $\mathbf{P}^{M,O}_t$ represent the points in timestep $t$ that belongs to either the tool or anchor object, $O\in{T,A}$, used for either demonstration or imitation $M\in{D,I}$. In DINOBot, only the first frame of the image trajectory is used, i.e., the bottleneck frame, assuming that the robot is positioned in a bottleneck pose that allows for the same interaction to be replayed from similar poses relative to novel objects. In this work, we add an additional preview pose that shows the anchor object with less occlusion before the bottleneck pose, to alleviate issues with the robot-held tool objects occluding the wrist-mounted camera.

\subsection{Overview}
The DINOBot framework consists of retrieving the most relevant demonstration, aligning with it, and replaying the demonstrated trajectory. The retrieval phase is not the focus of this work and is, thus, skipped. Instead, we present several innovations over the alignment phase.

The alignment phase establishes pixel-level correspondences between the anchor objects from the demonstration and the novel observed scene. The robot moves such that the spatial distribution of correspondences observed from the wrist camera is closer to the one in the demonstrated image. This is achieved by iteratively moving to the transformation which minimizes an error criterion (described in Section~\ref{method:error}). This process is repeated until the observed error is within some predetermined threshold, after which the replay phase begins.

In contrast to DINOBot, which employs global alignment, we restrict the matching to Regions of Interaction (RoIs) to ensure that only task-critical object parts contribute to the alignment. To robustly estimate the rigid transformation between the demonstration and the observed object, we integrate a RANSAC-based procedure to discard outliers. Finally, the computed transformation is used to reproject the demonstration trajectory into the observed object's coordinate frame for execution.

\subsection{Interaction-Focused Correspondence}
Global alignment may incorporate correspondences from regions that are irrelevant to the manipulation task. To improve functional relevance, we focus on Regions of Interaction (RoIs) that are critical for executing the task.

During demonstration, we take two RGB observations of the anchor object: one previewing the object and one at the bottleneck pose. The preview frame shows the anchor object without occluding it with the held tool. We find the point pairs between the anchor points from the preview frame $ \mathbf{P}^{D,A}_{preview}$ and the tool points at the bottleneck frame $ \mathbf{P}^{D,T}_{bottleneck}$ which are best buddies, i.e., mutual nearest neighbors. A distance threshold $d_{thresh}$ is applied originating from any point that was found to be within the pairs. The points belonging to the anchor object that are within the threshold can then be back-projected to the preview frame to produce a task-critical interaction region mask, or RoI.

Finally, during alignment, the DINO-ViT semantic features associated with the masked RoI are used for best-buddies correspondence with the DINO-ViT features in the novel anchor object. Best buddies matching is applied again, but on the semantic descriptors, producing the set of best-buddies pairs \( M \), defined as:
\begin{equation}
\begin{aligned}
    M=&\{(i,j) \mid p^{D}_i \in \mathbf{P}^{D,T}_{preview}, p^{I}_j\in \mathbf{P}^{I,T}_{0},
    \\&\text{ and } p^{D}_i \text{ and } p^{I}_j \text{ are mutual nearest neighbours}\}.
\end{aligned}
\end{equation}
The 2D anchor points in $M$ are then projected in 3D and expressed relative to the camera frame. A rigid transformation
\begin{equation}
    T^*=\arg\min_T \sum_{(i,j)\in M}||Tp^{D}_i-p^{I}_j||^2
\end{equation}
can be computed with SVD which aligns the robot's wrist camera relative to the new object's pose. This is done iteratively until the transformed points are close enough to their expected positions in the wrist camera's view.

\subsection{RANSAC-Based Outlier Rejection}
Dense pixel-level correspondences, even when computed with robust semantic descriptors, may include spurious matches due to occlusions, sensor noise, or non-discriminative regions. When focusing on the smaller task-critical RoI, there are naturally fewer matches than when performing full-object alignment. Thus, the optimization shown above is more sensitive to such noise. To mitigate these effects, we adopt a RANSAC (Random Sample Consensus) within the alignment phase:
\begin{enumerate}
    \item \textbf{Initial Matching:} For each pixel in the query image, the DINO-ViT features are extracted and matched to the corresponding features in the demonstration image using cosine similarity. This process yields a set of candidate point pairs, $\{(p^{I}_i, p^{D}_i)\}_{i=1}^{M}$.
    \item \textbf{Hypothesis Generation:} A random subset (typically 50\% of the candidate matches) is sampled to compute a candidate rigid transformation $T$:
    \begin{equation}
        T^* = \arg\min_T \sum_{i=1}^{N} \|T\, p^{I}_i - p^{D}_i\|^2,
    \end{equation}
    where $N$ is the number of sampled pairs. DINOBot originally used least squares optization based on SVD which can produce valid rotations but also occasionally reflections, which are not physically possible. Instead, here we use the Kabsch-Umeyama SVD variant~\cite{Kabsch1976,Umeyama1991} which is more robust and does not produce reflections.
    \item \textbf{Consensus Evaluation:} This procedure is repeated for a fixed number of iterations (e.g., $100$ iterations), and the transformation yielding the lowest average L2 distance error between the correspondence pairs is selected as the final estimate. This candidate transformation is then applied to all available correspondences.
\end{enumerate}
This RANSAC procedure effectively rejects outlier matches, ensuring that the estimated transformation is robust despite noisy and erroneous correspondences.

\subsection{Error Computation and Convergence Criterion}\label{method:error}
In the original DINOBot, the convergence of the alignment phase is determined by minimizing the overall distance
\begin{equation}
\mathcal{D} = \sqrt{\sum_{(i,j)\in M} \sum_{d\in{\{x,y,z\}}}
\left( \mathbf{p}^{D}_{i,d} - \mathbf{p}^{I}_{j,d} \right)^2}
\end{equation}
between the transformed point sets, where $d$ is the spatial dimension. However, such a metric is sensitive to the absolute number of correspondences. Although fewer matches may suffice for alignment, a greater number of high-quality matches typically improves precision. In practice, finding many reliable matches is challenging since many match candidates have to be discarded due to invalid or noisy depth, or corresponding between objects that have small or semantically different shapes. We instead adopt an average error metric defined as:
\begin{equation}
    \mathcal{D} = \frac{1}{|M|}
    \sum_{(i,j)\in M}
    \sqrt{
    \sum_{d\in{\{x,y,z\}}}
    \left( \mathbf{p}^{D}_{i,d} - \mathbf{p}^{I}_{j,d} \right)^2
    },
\end{equation}
where $|M|$ is the number of valid point pairs and the summation is performed over the three spatial dimensions. This metric normalizes the error by the number of correspondences, thereby providing a consistent convergence criterion irrespective of the point count. The alignment process iterates until $\mathcal{D}$ falls below a specified threshold (e.g. $0.004\,m$).

\subsection{Trajectory Replay}
After obtaining the optimal alignment transformation $T^*$, the recorded demonstration trajectory, represented as a series of end-effector poses $\{\mathbf{T}_t\}$, is re-projected into the coordinate frame of the observed object. The process involves:
\begin{enumerate}
    \item \textbf{Relative Trajectory Computation:} The demonstration trajectory is first expressed relative to the bottleneck pose:
    \begin{equation}
        \mathbf{T}_{\text{relative}} = T_{bottleneck}^{-1} \cdot \mathbf{T}_t.
    \end{equation}
    \item \textbf{Trajectory Transformation:} The relative trajectory is then transformed using the current observed pose:
    \begin{equation}
        \mathbf{T}^*_t = T_{\text{current}} \cdot \mathbf{T}_{\text{relative}},
    \end{equation}
    where $T_{\text{current}}$ is obtained from the alignment phase.
\end{enumerate}
This transformation ensures that the manipulation skill is faithfully reproduced on the novel object configuration.

\subsection{Implementation Details}
We employ the DINO-ViT-S model for extraction of semantic features and best-buddies matching for correspondence estimation. RANSAC is configured to sample 50\% of candidate matches over $100$ iterations, with an inlier threshold defined by an average error of $0.004\,m$. Invalid or noisy depth measurements are filtered using a statistical outlier removal algorithm. The Regions of Interaction are determined using masks generated by Grounded SAM~\cite{ren2024grounded}, and the optimal transformation is computed with the Kabsch-Umeyama algorithm to ensure proper rotations. This comprehensive approach ensures robust and precise alignment necessary for successful one-shot imitation in real-world robotic tasks.

\section{Experimental Evaluation}
In this section, we present a comprehensive evaluation of our proposed method. We assess performance on a diverse set of real-world manipulation tasks and conduct both quantitative and qualitative analyses. Our experiments are designed to evaluate the robustness, efficiency, and generalization of the improved alignment framework.

\subsection{Experimental Setup}
Our evaluation comprises two sets of experiments: single-task imitation and long-horizon task execution.

\textbf{Single-Task Imitation:}
We evaluate the method on seven distinct real-world manipulation tasks, including picking, insertion, shaking, and pouring. For each task, a single demonstration is provided using one set of objects, and the learned policy is tested on two to three novel variations of these objects. For every task-object combination, 10 independent trials are conducted.

\textbf{Long-Horizon Task Execution:}
To assess the robustness of the method in sequential operations, we design a composite task comprising five sequential subtasks: pick wine, pour wine into a container, pick coke, pour coke into a container, and shake the container. This task is repeated $10$ times with a fixed configuration, and the outcome of each subtask is recorded to evaluate cumulative performance.

\subsection{Evaluation Metrics}
The primary metrics used in our evaluation are:
\begin{itemize}
    \item \textbf{Success Rate:} Defined as the percentage of trials where the imitation accurately reproduces the demonstrated behavior. For long-horizon tasks, the overall success rate is computed as the fraction of successful runs at each stage over all attempts.
    \item \textbf{Alignment Efficiency:} Measured by the average number of alignment iterations required to achieve convergence, where convergence is defined as obtaining an average error $\mathcal{D}$ below a threshold (set at $0.004\,m$).
    \item \textbf{Failure Mode Analysis:} Trials are categorized based on failure types:
    \begin{enumerate}
        \item \textbf{Collision Failure during Alignment:} Predicted alignment poses that result in collisions.
        \item \textbf{Collision Failure during Replay:} Transformed trajectories that cause collisions during execution.
        \item \textbf{Semantic Failure:} Mismatches in the functional behavior of the demonstration, e.g., incorrect pouring or grasping locations.
        \item \textbf{Convergence Failure (Timeout):} Inability to achieve a sufficiently low alignment error within $10$ iterations.
    \end{enumerate}
\end{itemize}

\subsection{Quantitative Results}
\begin{figure}[t]
    \centering
    \includegraphics[width=\linewidth]{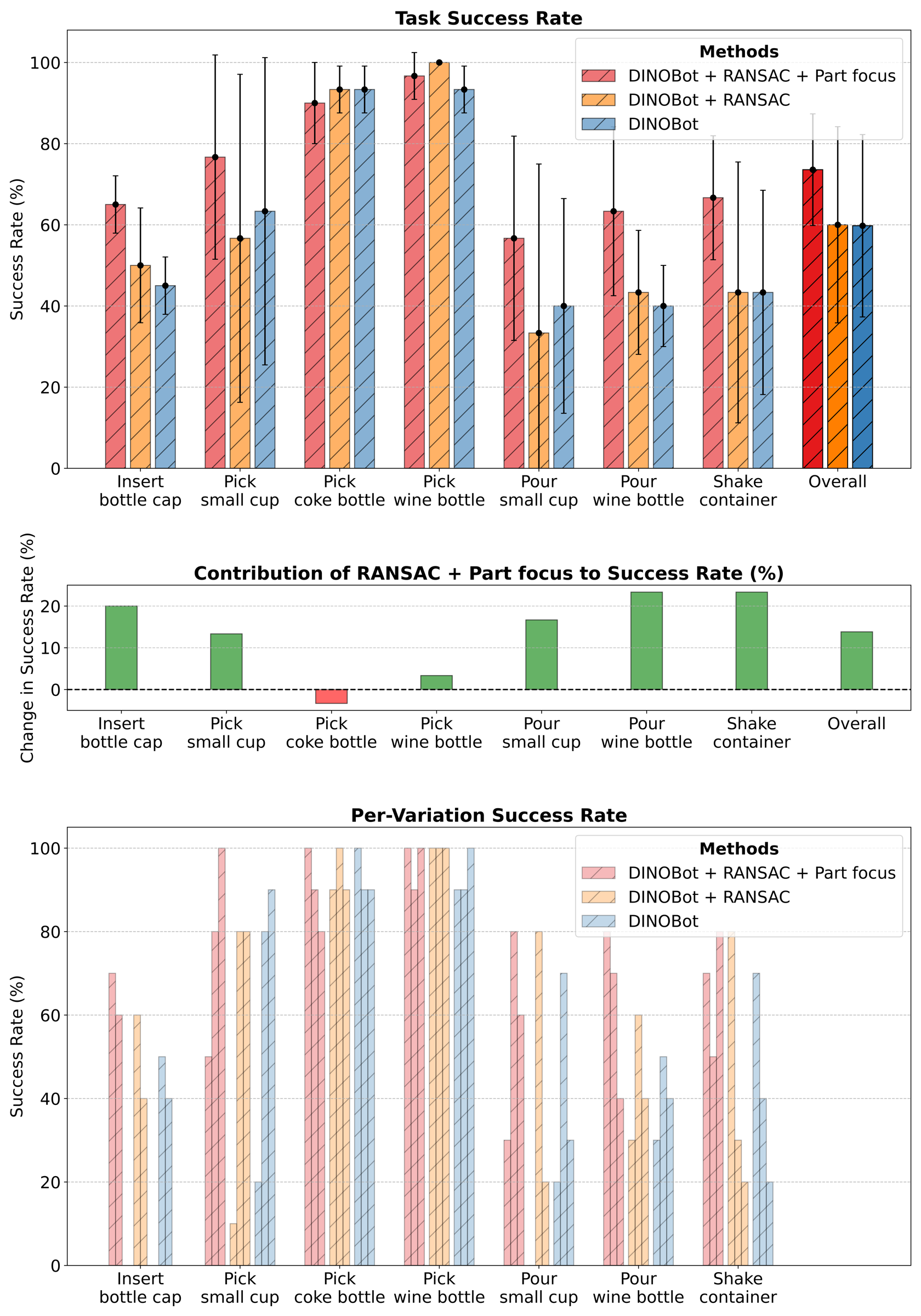}
    \caption{Task success rates for the proposed method versus the baseline. Our approach outperforms the baseline by 13.8\% overall.}
    \label{fig:affimit:success}
\end{figure}

Figure~\ref{fig:affimit:success} presents a comparison of task success rates between our proposed method and the baseline DINOBot framework. Our approach achieves an overall improvement of 13.8\% in the success rate. Statistical significance was confirmed using the Wilcoxon signed-rank test ($W=3.5$, $p=0.00296$).

\begin{figure}[t]
    \centering
    \includegraphics[width=\linewidth]{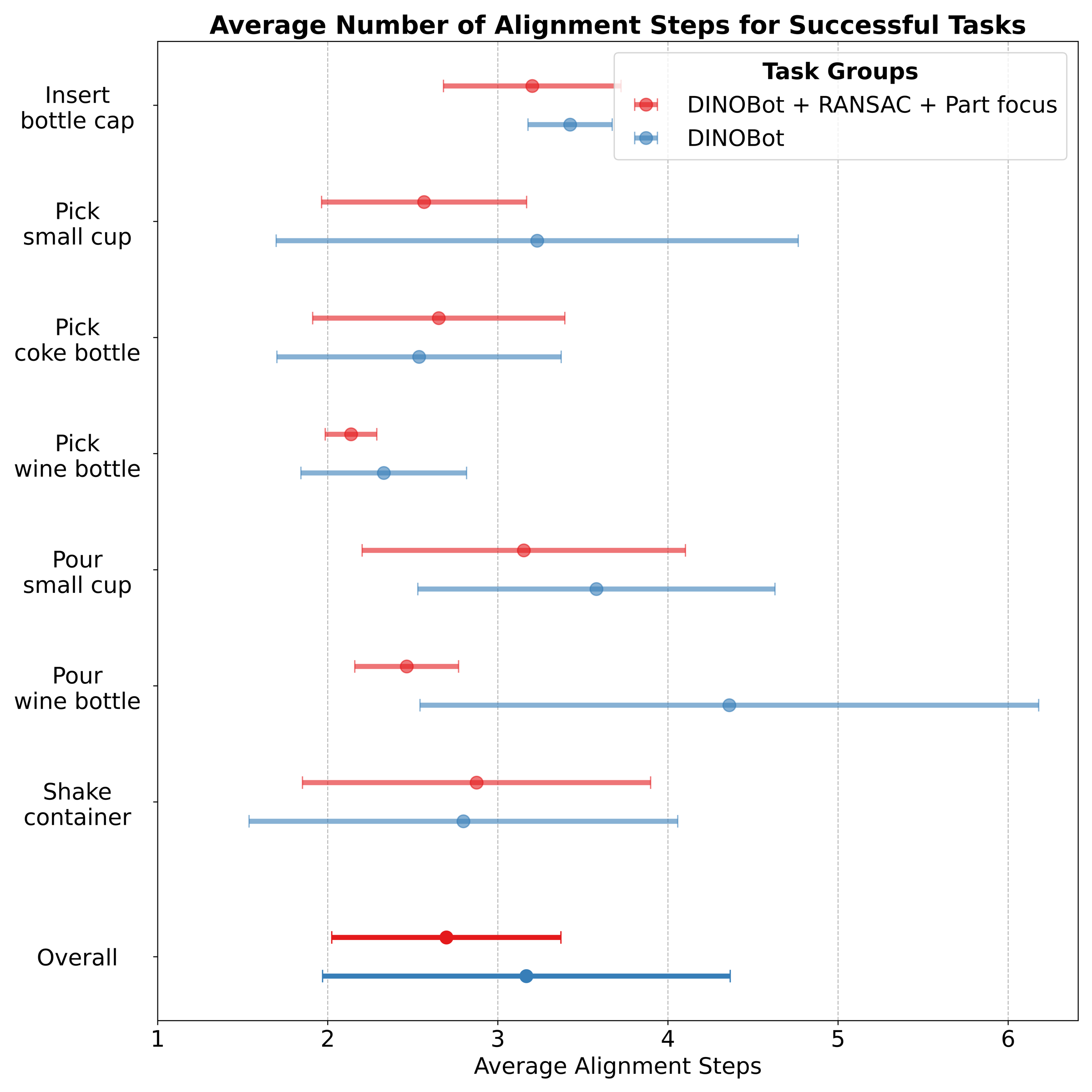}
    \caption{Average number of alignment steps needed for convergence. Our method converges faster than the baseline, indicating improved robustness.}
    \label{fig:affimit:avgsteps}
\end{figure}

Figure~\ref{fig:affimit:avgsteps} illustrates the average number of alignment iterations required for successful convergence. The proposed modifications reduce the average iteration count, demonstrating increased computational efficiency and robustness in aligning challenging object pairs.

\subsection{Qualitative Analysis and Failure Breakdown}
\begin{figure}[t]
    \centering
    \includegraphics[width=\linewidth]{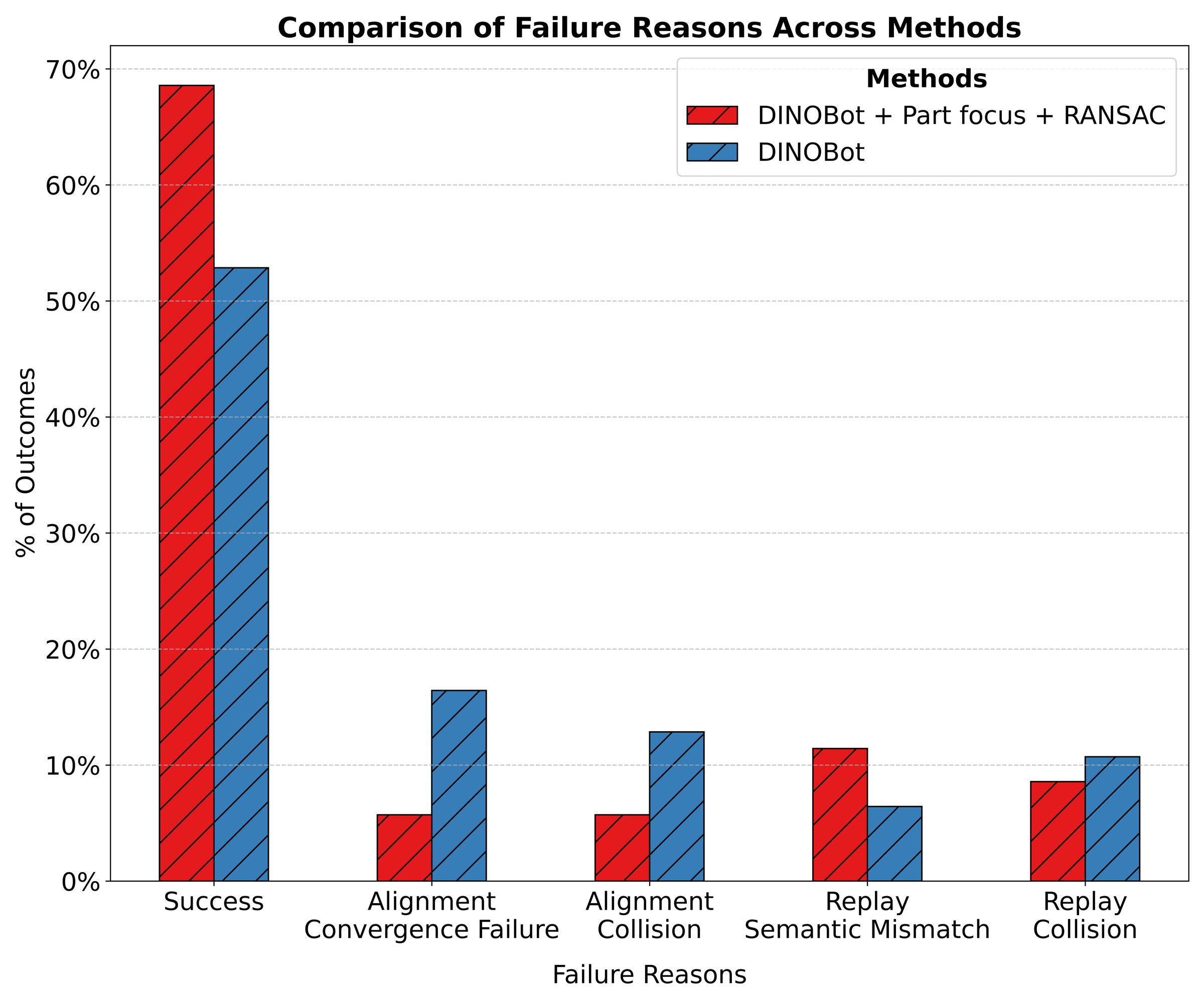}
    \caption{Breakdown of failure modes for both the baseline and the proposed method. The enhanced method notably reduces convergence failures, with some increase in semantic mismatches under challenging conditions.}
    \label{fig:affimit:single-failures}
\end{figure}

A detailed failure mode analysis (see Figure~\ref{fig:affimit:single-failures}) reveals that our method substantially reduces convergence failures and the number of collisions. However, an increase in semantic failures was observed in scenarios involving reflective objects that introduce significant depth noise. These results suggest that while our method improves alignment reliability, further sensor noise mitigation could enhance performance even further.

\subsection{Long-Horizon Task Execution}
\begin{figure}[t]
    \centering
    \includegraphics[width=\linewidth]{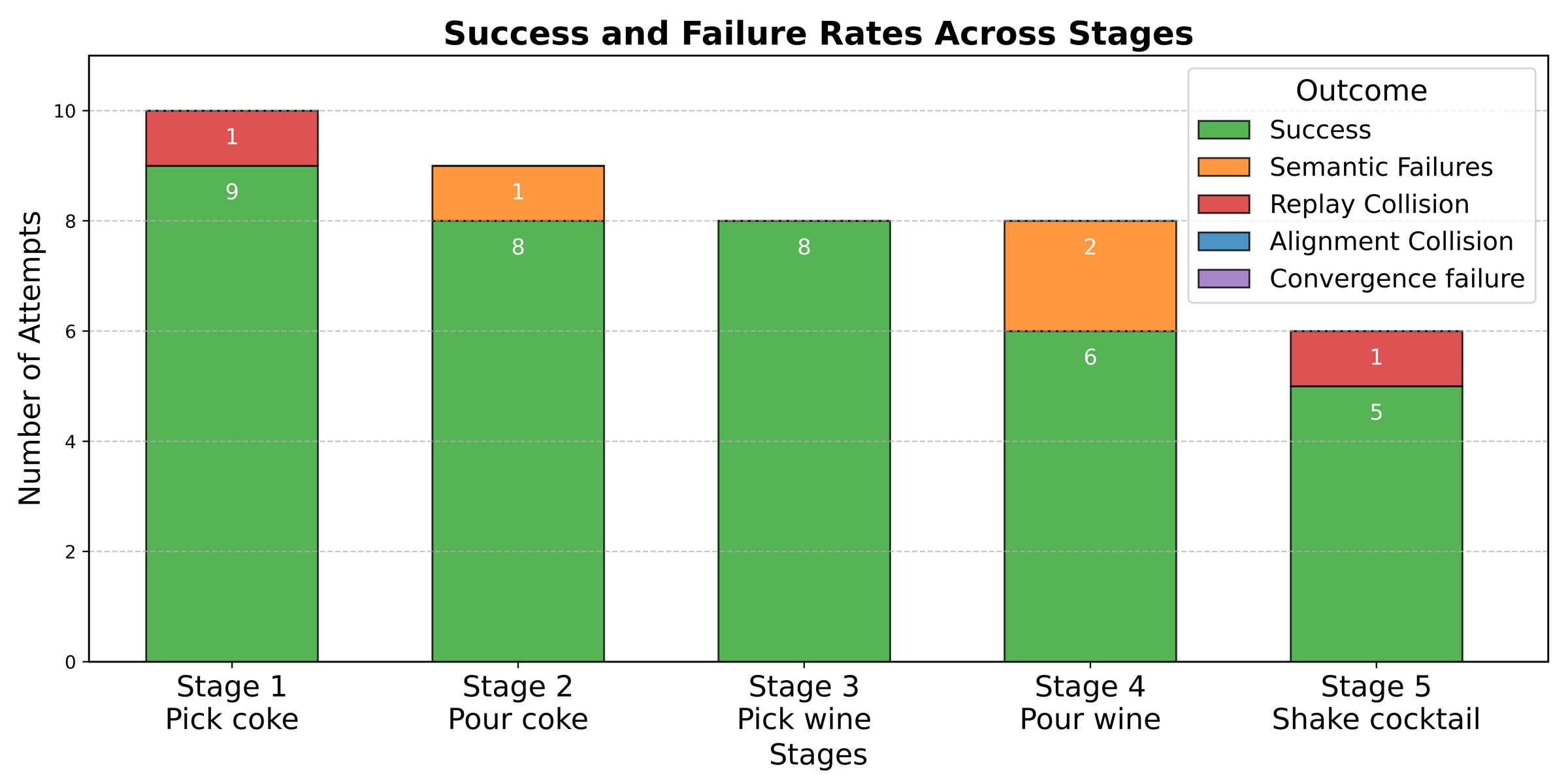}
    \caption{Long-horizon task execution results. The system completed the five-stage task in 5 out of 10 trials, matching the expected performance from individual task success probabilities.}
    \label{fig:longhorizon}
\end{figure}

To evaluate the method's performance in extended operations, we executed a long-horizon task comprising five sequential subtasks. Figure~\ref{fig:longhorizon} summarizes the success rates for the composite task in 10 trials, showing a 50\% success rate, which closely aligns with the theoretical product of individual task success rates (approximately 51.2\%). This result indicates that the proposed approach maintains its performance in sequential stages with minimal error propagation.

\begin{figure}[t]
    \centering
    \includegraphics[width=0.9\linewidth]{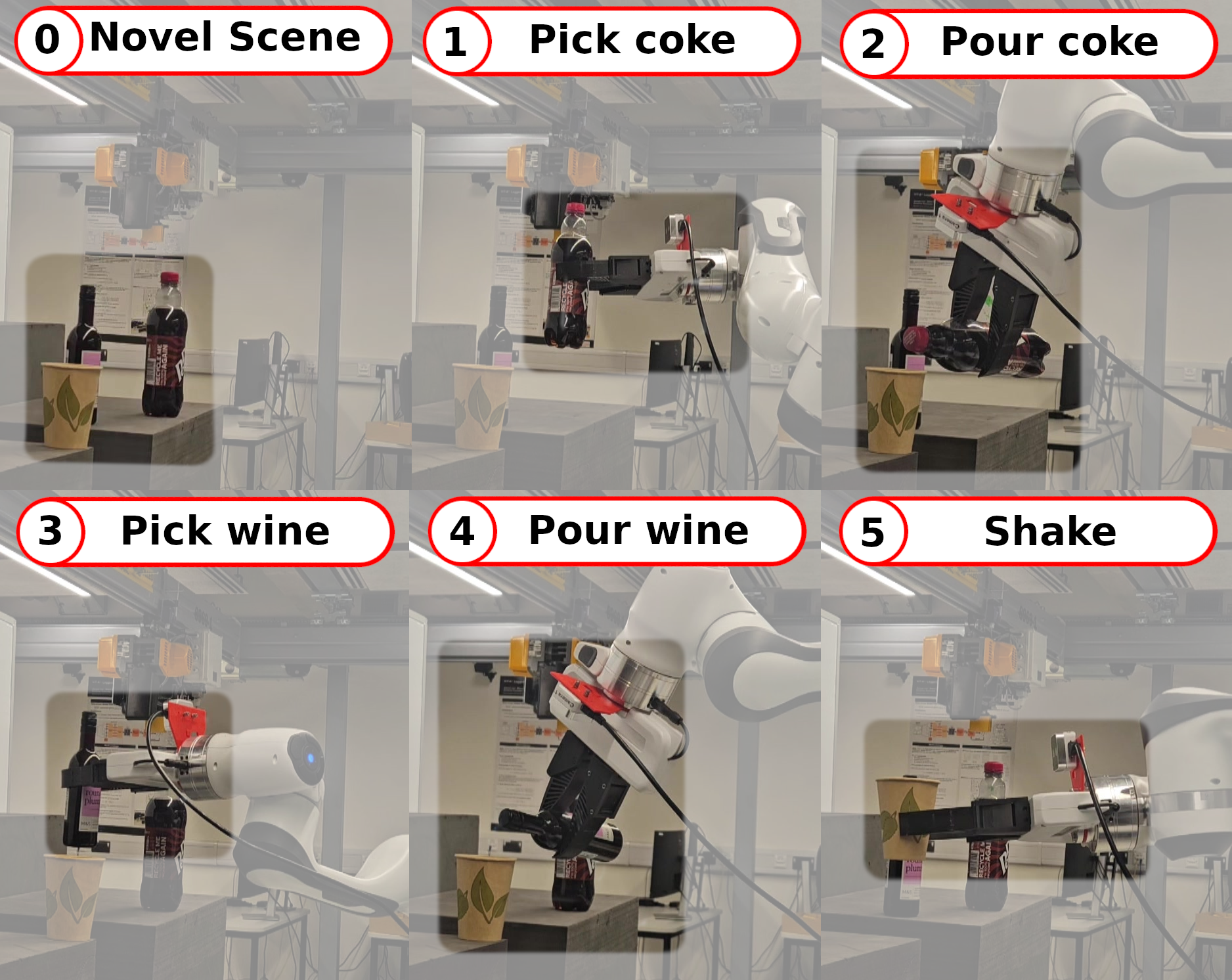}
    \caption{A successful long-horizon task execution. The robot demonstrates a seamless transition across the sequential manipulation subtasks.}
    \label{fig:affimit:composite}
\end{figure}

Figure~\ref{fig:affimit:composite} provides a visual example of a successful long-horizon execution, illustrating the smooth transition between subtasks and the effective transfer of the demonstrated trajectory.

\subsection{Discussion}
Quantitative and qualitative analysis confirm that our integration of outlier rejection and interaction-focused correspondence significantly enhances alignment accuracy and task performance. Reduced alignment iterations indicate improved efficiency but also greater reliability in the presence of sensor noise and object variability. Although the increase in the number of semantic failures highlights an area for future improvement, our method demonstrates robust one-shot imitation across a range of real-world scenarios. These results show the potential of our approach for deployment in more complex robotic manipulation tasks.

\section{Conclusion}
We presented a method for one-shot imitation that improves on DINOBot, significantly enhancing manipulation skill transfer. Our approach improves upon DINOBot by focusing alignment on task-critical Regions of Interaction and mitigating the adverse effects of noisy depth data and spurious feature correspondences using RANSAC outlier rejection. Quantitative experiments on various real-world tasks demonstrate an overall success rate improvement of 13.8\% over the baseline DINOBot, accompanied by a reduction in the average number of alignment iterations. Furthermore, evaluations of long-horizon tasks confirm that our method maintains reliable performance across sequential manipulation stages, indicating its suitability for complex robotic applications.

Despite these promising results, certain limitations remain. Our failure mode analysis reveals that while convergence failures are reduced, semantic mismatches persist, particularly in scenarios with reflective or low-contrast surfaces. These challenges suggest that further improvements in sensor data pre-processing and the robustness of semantic descriptors are necessary to fully address ambiguities in object interaction.

Future work will focus on several key areas:
\begin{itemize}
    \item \textbf{Point Cloud Completion:} Investigating advanced approaches for point cloud completion to improve the quality of depth data in challenging environments, e.g. by representing the scene via neural radiance fields.
    \item \textbf{Cross-Embodiment Imitation:} Extending the framework to support demonstrations from diverse embodiments, including human-to-robot skill transfer, which would broaden the applicability of the method across different robotic platforms.
    \item \textbf{Context-Aware Planning:} Adapting the skills used to the context of long-horizon tasks, for example, adapting the grasping strategy for a bottle with the intention of pouring as opposed to handing it over to another robot.
\end{itemize}

In summary, our work marks a significant step towards more robust and precise one-shot imitation for robotic manipulation.

\bibliographystyle{IEEEtran}
\bibliography{ref}

\end{document}